\documentclass{fairastra}

\usepackage[T1]{fontenc}
\usepackage{lmodern}
\usepackage{multirow}
\usepackage{soul}
\usepackage{pifont}
\usepackage[T1]{fontenc}

\usepackage{graphicx}
\usepackage{amsmath,amssymb}

\usepackage{makecell}
\usepackage[dvipsnames]{xcolor}
\usepackage{color, colortbl}

\usepackage{caption}
\usepackage{algorithm}

\usepackage{xspace}
\makeatletter
\DeclareRobustCommand\onedot{\futurelet\@let@token\@onedot}
\def\@onedot{\ifx\@let@token.\else.\null\fi\xspace}

\makeatother

\usepackage{newtxtext}

\usepackage{algpseudocode}
\definecolor{catgray}{gray}{0.92}

\newcommand{\emailsym}{\Letter}

\usepackage[dvipsnames]{xcolor} 
\definecolor{FutureOrange}{HTML}{EC866D}
\usepackage{marvosym}
\usepackage{fontawesome5}
\title{OmniNav:A Unified Framework for Prospective Exploration and Visual-Language Navigation}
\author{\parbox{\linewidth}{
 Xinda Xue$^{1,2*}$\quad Junjun Hu$^{1*\dagger{\textsuperscript{\emailsym}}}$\quad Minghua Luo$^{1*}$\quad Shichao Xie $^{1}$\quad Jintao Chen$^{1,2}$ \\
 \textbf{Zixun Xie}$^{2}$\quad \textbf{Kuichen Quan}$^{1}$\quad \textbf{Wei Guo}$^{1}$\quad \textbf{Mu Xu}$^{1}$\quad \textbf{Zedong Chu}$^{1}$ \\[0.5em]
$^{1}$Amap, Alibaba Group \quad
$^{2}$Peking University \\[0.5em]
\small\mdseries
\texttt{\{xuexinda.xxd, hujunjun.hjj, luominghua.lmh, tenan.xsc, anyi.cjt, quankuichen.qkc, weisheng.gw, xumu.xm, chuzedong.czd\}@alibaba-inc.com}
}
}
\contribution[*]{Equal Contribution}
\contribution[\dagger]{Project Lead.}
\contribution[]{\raisebox{-1pt}{\textsuperscript{\emailsym}}Corresponding authors}

\abstract{
Embodied navigation is a foundational challenge for intelligent robots, demanding the ability to comprehend visual environments, follow natural language instructions, and explore autonomously. However, existing models struggle to provide a unified solution across heterogeneous navigation paradigms, often yielding low success rates and limited generalization. We present OmniNav, a unified framework that handles instruct-goal, object-goal, point-goal navigation, and frontier-based exploration within a single architecture. First, we introduce a lightweight, low-latency policy that predicts continuous-space waypoints (coordinates and orientations) with high accuracy, outperforming action-chunk methods in precision and supporting real-world deployment with control frequencies up to 5 Hz. Second, at the architectural level, OmniNav proposes a fast-slow system design: a fast module performs waypoint generation from relatively short-horizon visual context and subtasks, while a slow module conducts deliberative planning using long-horizon observations and candidate frontiers to select the next subgoal and subtask. This collaboration improves path efficiency and maintains trajectory coherence in exploration and memory-intensive settings. Notably, we find that the primary bottleneck lies not in navigation policy learning per se, but in robust understanding of general instructions and objects. To enhance generalization, we incorporate large-scale general-purpose training datasets including those used for image captioning and referring/grounding into a joint multi-task regimen, which substantially boosts success rates and robustness. Extensive experiments demonstrate state-of-the-art performance across diverse navigation benchmarks, and real-world deployment further validates the approach. OmniNav offers practical insights for embodied navigation and points to a scalable path toward versatile, highly generalizable robotic intelligence.

}

\date{Jan. 6, 2026}
\astradata[Project Page]{\url{https://astra-amap.github.io/omninav.github.io/}}
\astradata[Github]{\url{https://github.com/amap-cvlab/OmniNav/}}

\begin{document}

\maketitle

\section{INTRODUCTION}

Embodied navigation~\citep{gao2024vision,gu2022vision} has emerged as a core problem in embodied intelligence: enabling robots to perceive, understand, and explore real-world environments without pre-built maps while following natural language instructions. To act reliably in dynamic, partially observable environments, an agent must not only ground instantaneous visual inputs but also maintain coherent spatiotemporal memory and perform active exploration. Application demands for real-time responsiveness further increase the requirements for low-latency decision-making and cross-environment generalization.

Current research largely revolves around three paradigms: point-goal~\citep{liu2025citywalker}, instruct-goal~\citep{anderson2018vision,ku2020room}, and object-goal~\citep{yokoyama2024hm3d}. point-goal tasks are well-specified and straightforward to evaluate but rely on explicit coordinates rarely available in practice; instruction-goal aligns with human usage but often generalizes poorly to unseen instructions or environments; object-goal is the most practical but requires robust target recognition coupled with efficient path planning, making it the most challenging. Many existing methods remain customized, relying on task-specific data, which limits cross-task transfer and the potential for mutual enhancement. Uni-Navid~\citep{zhang2024uni} proposed a VLM-based discrete action predictor unifying vision-and-language navigation, object-goal navigation, embodied question answering~\citep{das2018embodied}, and following~\citep{wang2025trackvla}, but its study of LLM long-horizon planning is not sufficiently developed. MTU3D~\citep{zhu2025move} advances a “move to understand” paradigm by coupling frontier exploration with visual localization in a single objective, yet requires constructing 3D object coordinates, leading to deployment complexity. Although recent Video-LLMs~\citep{wei2025streamvln,qi2025vln,zhang2024navid} and VLAs~\citep{sapkota2025vision,zitkovich2023rt,ma2024survey} integrate vision, language, and action prediction end to end, they still face bottlenecks in streaming video input, long-context management, and low-latency inference: discretized action modeling sacrifices precision and flexibility; constrained LLM call frequency and frequent context resets lead to deployment difficulties; besides, in practice, the dominant failure mode often stems from inadequate understanding of generic instructions and open-vocabulary objects rather than policy learning itself. These gaps call for a unified, efficient framework that balances long/short-horizon reasoning with real-time responsiveness.

We present OmniNav, a unified embodied navigation framework that concurrently covers instruct-goal, object-goal, point-goal, and frontier-based exploration within a single architecture. Inspired by dual-system theory~\citep{figure2024helix,black2025pi0}, OmniNav coordinates a fast–slow system~\citep{black2025pi0}: a fast system reacts to comparatively short-horizon perception and current tasks or subtasks, generating high-precision waypoints (coordinates and orientations) to support low-latency control up to 5 Hz; a slow system deliberates over long-horizon observations and frontier cues, leveraging a VLM’s chain-of-thought~\citep{wei2022chain} to decompose complex goals and select the next subgoal and subtask. The two are coupled through a central memory module that uses a key–value (KV) cache to provide essential spatiotemporal context, yielding decisions that are both locally agile and globally consistent.

OmniNav addresses the triad of real-time operation, fast–slow collaboration, and generalization. A lightweight flow-matching policy~\citep{bjorck2025gr00t} avoids the precision degradation and latency accumulation inherent to action discretization; fast–slow collaboration ensures exploration efficiency and trajectory coherence in long-memory scenarios; more importantly, training unifies large-scale generic vision–language data (captioning, referring/grounding, etc.) with multiple navigation tasks, significantly strengthening instruction following and open-vocabulary object perception to improve success rates and robustness. Our contributions are threefold:
\begin{itemize}
\item A unified architecture that, under a single training framework and policy, supports multiple goal modalities (point, object, and instruction) as well as frontier-based exploration;
\item An end-to-end fast–slow coordination with central memory that reconciles low-latency control and high-level deliberation;
\item A principled strategy to incorporate generic vision–language data into joint training, systematically improving cross-task and cross-environment generalization. 
\end{itemize}
Extensive experiments set new state-of-the-art results across multiple navigation benchmarks, with real-robot deployments further validating practicality. We contend that OmniNav charts a scalable path toward multifunctional, highly generalizable embodied navigation systems.

\section{RELATED WORKS}
\label{headings}

\textbf{Vision Language Models for Navigation}
Leveraging their powerful generalization capabilities in understanding and planning, Visual Language Models (VLMs)~\citep{chiang2023vicuna,liu2023visual,zhu2023chatgpt} have been increasingly applied to the domain of robotic navigation, achieving notable success. Prevailing methods~\citep{dorbala2022clip,zhou2024navgpt,long2024discuss} typically employ VLMs to process multimodal instructions and directly decode low-level actions in an autoregressive manner. However, this paradigm suffers from significant drawbacks: it is prone to compounding errors in sequential prediction and is often hampered by slow inference speeds. In contrast, our approach draws inspiration from recent advances in Vision-Language-Action (VLA) models~\citep{zitkovich2023rt,kim2024openvla,li2024manipllm}. We introduce a novel architecture that appends a flow-matching policy~\citep{zhao2024monoformer,chen2024diffusion,zhou2024transfusion} to a VLM backbone. This design enables our model to generate entire action trajectories non-autoregressively, leading to substantially improved prediction accuracy and computational efficiency, especially when navigating unseen environments.

\textbf{Dual-System Design}
Dual-system architectures have been widely adopted across various domains to meet diverse operational demands. In the realm of Vision-Language-Action (VLA) models, several works~\citep{bjorck2025gr00t,bu2025agibot,ge2024seed,song2025hume} have implemented dual-system designs to balance fast control execution and intelligent planning. Inspired by this paradigm and motivated by the specific requirements of embodied navigation, we propose a novel dual-system framework.
Our framework consists of two complementary components. The first is a fast system, a purely visual, end-to-end policy designed for direct deployment and is highly effective in the majority of navigation scenarios. The second system is specifically engineered for challenging long-horizon tasks. To serve as a long-term memory mechanism, we employ a planning strategy that combines frontier-based exploration~\citep{zhu2025move} with images. This approach offers a more concise implementation compared to alternative memory structures such as scene graphs~\citep{team2025robobrain} or complex semantic maps~\citep{long2024instructnav}. These alternatives memory structures can also be implementations for the slow system. The idea stays the same: the slow system is responsible for global planning, while the fast system handles local execution. This synergistic design has proven its superiority by achieving State-of-the-Art performance on multiple benchmarks.

\textbf{Frontier-based Navigation}
Recent studies on exploration and navigation adopt different strategies for selecting informative targets in unknown environments. GOAT~\citep{chang2023goat} and its benchmark GOAT-Bench~\citep{khanna2024goat} study lifelong navigation and object search using an object-instance memory and frontier exploration. Similarly, MTU3D~\citep{zhu2025move} keep an object-goal memory built from 3D point clouds and semantic segmentation, and combine this with frontier exploration. A different group of methods uses non-semantic frontier exploration~\citep{chang2023goat,sakamoto2024map,nayak2025multi}, where the next target is usually just the closest frontier, sometimes adjusted by simple heuristics such as distance–heading scores. OmniNav instead uses a semantics- and reasoning-aware frontier selection: it links each frontier to its egocentric images, then uses explicit chain-of-thought reasoning over these views to decide which frontier is more informative or promising for the current task.

\section{Approach}
\label{others}

\begin{figure}[h] 
\centering 
\includegraphics[width=1.0\linewidth]{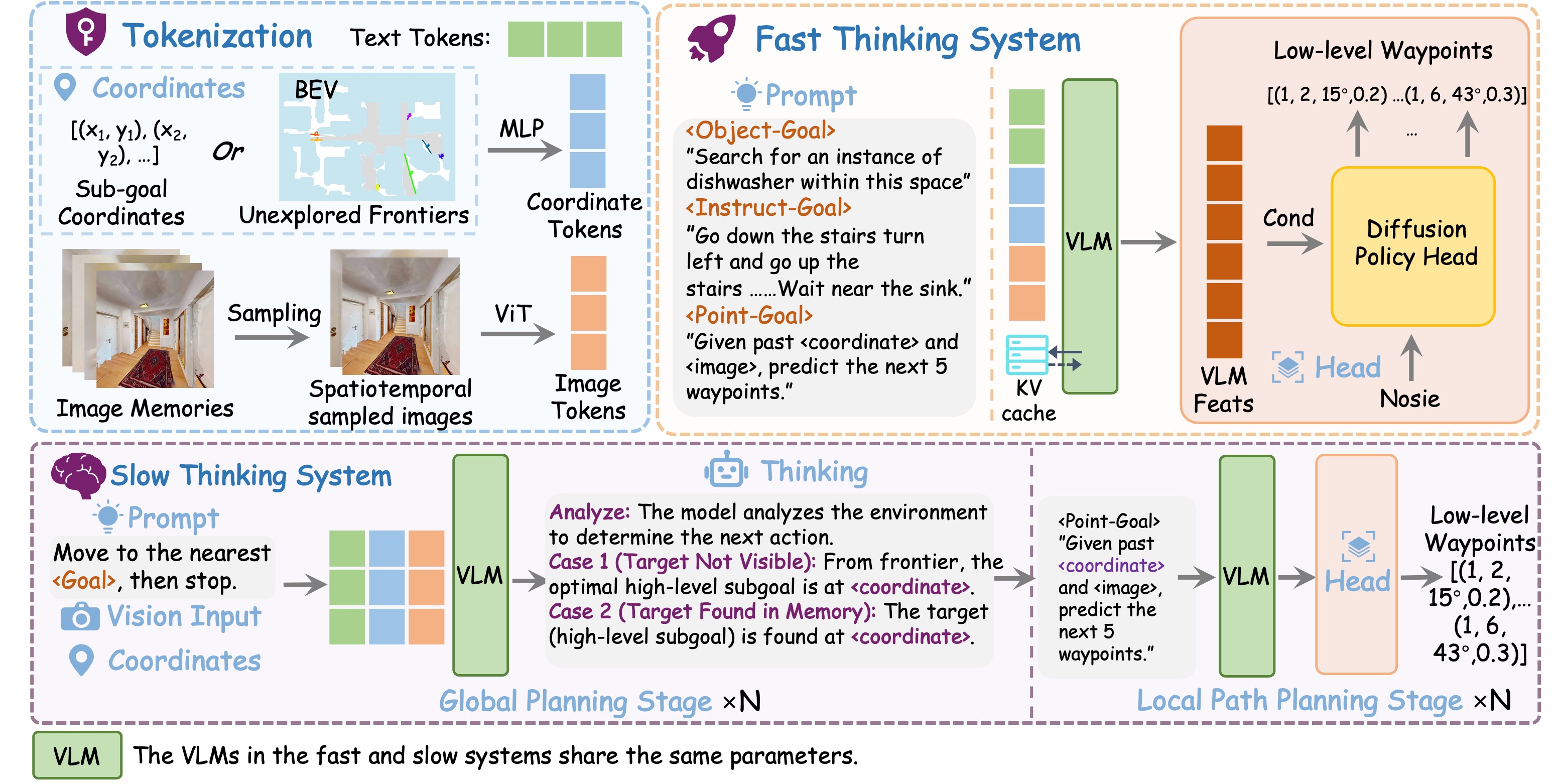}
\caption{The fast system can independently handle multi-task navigation, using the VLM backbone and a flow-matching policy to rapidly generate waypoints. Building on this, a slow thinking module is integrated to enable long-term memory and planning: it constructs long-range spatial and semantic memory using frontiers and images, and provides subgoal cues. The collaboration between the slow and fast proceeds as follows: the slow system uses frontiers or memory to generate high-level subgoals, once a subgoal is determined, the fast system takes over and progressively produces low-level waypoint sequences, ultimately reaching the target.}

\label{fig:model} 
\end{figure}

\textbf{Multimodal Input tokenizations}
To handle all four task types through a unified interface, text, coordinates, and visual history are converted into a set of discrete tokens consumable by a Large-Language Model (LLM) see Fig. \ref{fig:model}. We use Qwen2.5-VL-3B-Instruct~\citep{bai2025qwen2} as the base model and extend it with a coordinate modality. During streaming inference, a key–value (KV) cache is maintained to reduce latency.
Text tokens: Derived from natural-language task descriptions, object category labels, and point-goal commands, are all converted into a standardized instruction sequence. Coordinate tokens: Candidate search regions are represented as sets of 2D coordinates and heading angles, sourced from point-goal inputs or subgoal positions generated by the slow system. These coordinates are processed via an MLP to dense embeddings that serve as coordinate tokens.
Image tokens: The central memory maintains a ring buffer of pose-stamped images. For the fast system, it spatiotemporally samples from the historical image sequence, maintaining a maximum number of images (e.g., 20 frames). For the slow system, it samples images from the spatiotemporal neighborhood of candidate frontiers. All images are encoded with a ViT to produce image tokens.

\textbf{Fast Thinking System}
OmniNav operates at a high frequency, designed to execute either subtasks provided by a slow system or end-to-end multi-task navigation, as shown in Fig.~\ref{fig:model}. It parallelly outputs a sequence of 5 continuous-space waypoints,  $ \mathbf{w}_{t:t+H} \in \mathbb{R}^{H \times 5} $ with $ H = 5 $. We formulate waypoint prediction as a conditional diffusion generation task. The input coordinates are first embedded by an MLP and then encoded together with the images and texts by the VLM. The VLM performs deep fusion over these features, and the resulting fused features are used as conditions for the diffusion model, guiding waypoint generation. This design preserves the VLM’s semantic understanding while enabling rich interactions between language-guided context and the waypoint, leading to robust instruction following. Compared with conventional autoregressive methods, the policy head achieves an speedup and, with a history of 20 frames, supports an inference rate of 5 Hz for real-time closed-loop control. At the same time, it produces smoother and more precise trajectories.

We employ a variant of the Denoising Transformer (DiT)~\citep{peebles2023scalable} to model waypoint sequences. The policy network consists of self-attention blocks that operate on noised tokens to capture temporal and spatial dependencies within the waypoint sequence, and cross-attention blocks that attend to the vision-language context $\mathbf{O}_{VLM}$.
The output is a sequence of $H=5$ spatial-temporal waypoints $\mathbf{w}_t^{(i)} \in \mathbb{R}^5$, $i = 1, \dots, H$, each encoding:
\begin{equation}
    \mathbf{w}_t^{(i)} = \left( x^{(i)}, y^{(i)}, \sin \theta^{(i)}, \cos \theta^{(i)}, c^{(i)} \right),
\end{equation}
where $(x^{(i)}, y^{(i)})$ denotes the 2D position, $\theta^{(i)}$ is the orientation (represented via sine-cosine embedding to avoid discontinuity at $\pi/-\pi$), and $c^{(i)} \in \{0,1\}$ is a binary completion flag indicating whether the ``arrive'' command should be triggered at the $i$-th waypoint.

Conditional flow matching policy is employed~\citep{lipman2022flow}. Given a ground-truth waypoint sequence $\mathbf{w}_{t:t+H}$, noise $\boldsymbol{\epsilon} \sim \mathcal{N}(\mathbf{0}, \mathbf{I})$, and a time parameter $\tau \in [0,1]$, the input is constructed as:
\begin{equation}
    \mathbf{w}_{t:t+H}^\tau = \tau \mathbf{w}_{t:t+H} + (1 - \tau) \boldsymbol{\epsilon},
    \label{eq:noising_process}
\end{equation}
and the policy $\pi$ is trained to estimate the denoising residual $\boldsymbol{\epsilon} - \mathbf{w}_{t:t+H}$ by minimizing:
\begin{equation}
    \mathbb{E}_{\tau, \boldsymbol{\epsilon}} \left[ \left\| \pi(\mathbf{O}_{VLM}, \mathbf{w}_{t:t+H}^\tau) - (\boldsymbol{\epsilon} - \mathbf{w}_{t:t+H}) \right\|^2 \right].
    \label{eq:flow_matching_loss}
\end{equation}
At inference, waypoints are generated via $S=5$ steps of Euler integration. Starting from initial noise $\mathbf{w}_{t:t+H}^0 \sim \mathcal{N}(\mathbf{0}, \mathbf{I})$, we iteratively refine the sequence:
\begin{equation}
    \mathbf{w}_{t:t+H}^{\tau + \Delta\tau} = \mathbf{w}_{t:t+H}^{\tau} + \frac{1}{S} \pi(\mathbf{O}_{VLM}, \mathbf{w}_{t:t+H}^{\tau}), \quad \Delta\tau = \frac{1}{S},
    \label{eq:euler_update}
\end{equation}
with $\tau$ increasing from 0 to 1. The final denoised output $\mathbf{w}_{t:t+H}^1$ serves as the predicted waypoints.

\textbf{Slow Thinking System}
The slow thinking system is the deliberative planning module for hierarchical active exploration. Its core responsibilities are twofold: when the target appears in the current or historical field of view, it can quickly localize the target and generate the coordinates of the subgoal that drive the fast system to progressively approach it, when the target is not observed, it selects subgoal position with strong semantic relevance to the target to explore next. This process demands both exploration and environmental understanding. The former requires the model to navigate and discover the environment, while the latter involves a core task: predicting the spatial coordinates of the target based on input camera poses and images, this grounds the model's semantic understanding in a concrete, geometric output.

In the fast–slow system collaboration, the fast system is not just a low-level controller following preset coordinates or smoothing a pre-planned path. It must constantly use raw visual input to move toward its subgoal. For example, if it gets a coordinate from the slow system but a wall blocks the straight-line path, the fast system must use visual cues to find a path that avoids the obstacle. If it gets a target coordinate from object memory, it can adjust its final pose based on what it currently sees, such as stopping precisely at the left, right, or center of the object. By continuously updating and refining its waypoints while moving, the fast system can reach targets more accurately.

Frontier~\citep{zhu2025move} is employed to guide the active exploration. We maintain a 3D occupancy map, which categorizes each region as explored or unknown, and frontiers are then identified as the boundary points between explored and unknown regions. In addition, to comprehend past temporal and spatial information, we construct a memory bank~\citep{zhu2025move,olton1977spatial,xu2024embodiedsam,zhou2023dual}. This repository archives a history of observations, storing the visual data and corresponding pose information (coordinates and orientations) after every executed action. We then design a sampling strategy that connects this historical context to future exploration by collecting all historical images captured near the agent's current location. It then evaluates each frontier by iterating through these images, sampling the one whose original capture viewpoint is most suitably aligned with the frontier's spatial coordinates. This image thereby becomes a visual proxy for that frontier. 

\begin{figure}[h] 
\centering 
\includegraphics[width=1.0\linewidth]{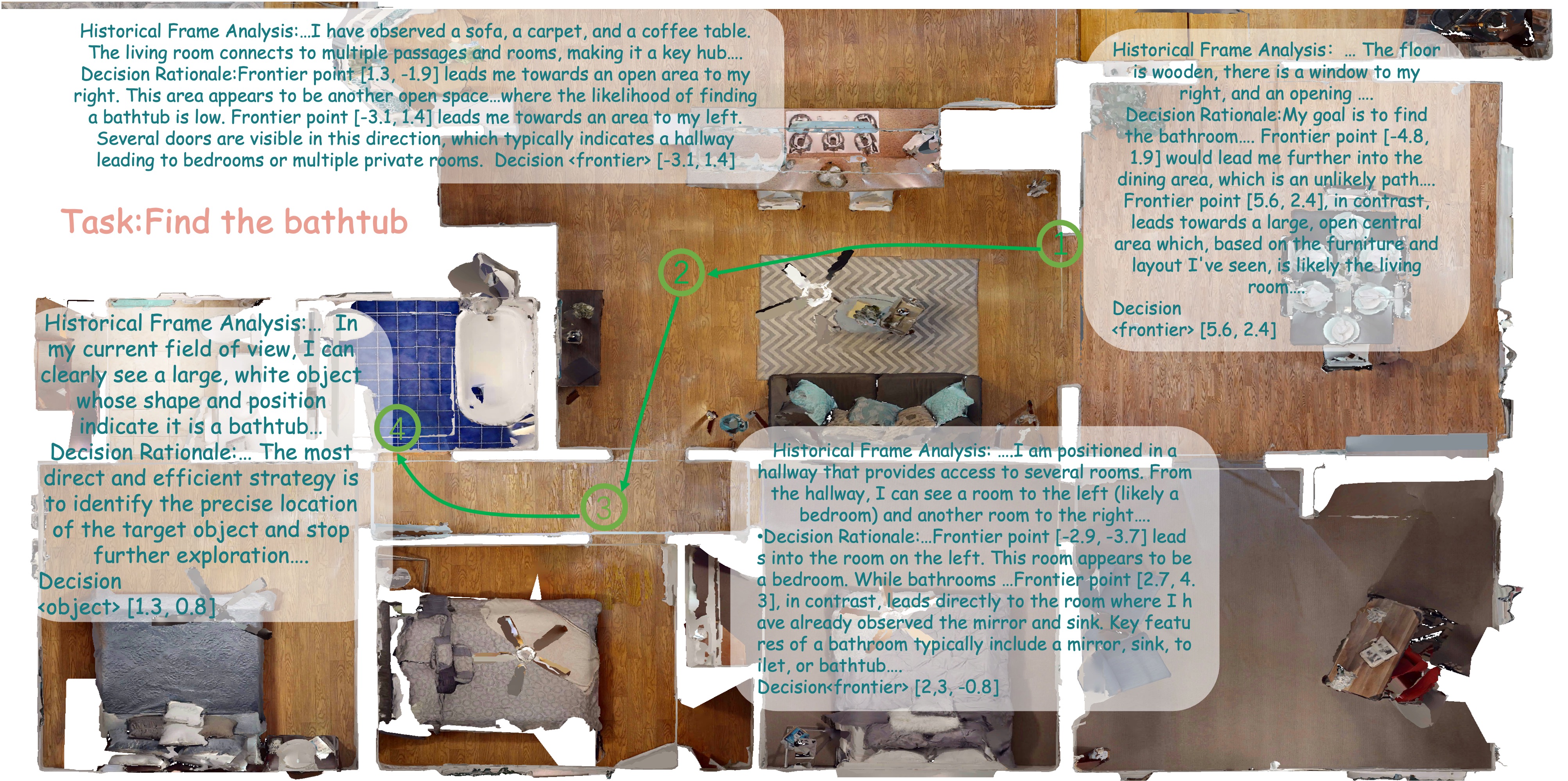}
\caption{Reasoning process by the slow system for exploration. For the “find the bathtub” task, the model reasons over the frontier set using memory and semantic priors, iteratively generating subgoals for the next exploration.}
\label{fig:cot} 
\end{figure}

During frontier selection, the model engages in comprehensive spatial and content reasoning, reflecting its capability to actively explore unknown environments and propose subgoal locations related to the target object. For example, when searching for a toilet, it prioritizes exploration locations in the bathroom; when searching for a television, it seeks locations associated with the living room. Moreover, if the target object appears in memory or within the current view, it outputs the target’s existing location.
We incorporate explicit chain-of-thought (CoT)~\citep{lin2025navcot,wang2025think,zhao2025cot} reasoning into the slow system’s prediction process to enable transparent process expression, achieving interpretability and self-correction. This also allows for richer textual outputs that strengthen the model’s grasp of logic and improve complex reasoning performance. Fig.\ref{fig:cot} illustrates the slow system’s reasoning process.

\section{DATA and Training}
\label{others}
Data in the embodied domain is typically organized as a data pyramid~\citep{bjorck2025gr00t}, with internet data and human video data at the bottom, simulation or synthetic data in the middle, and real-robot data at the top. Our dataset follows this rich composition as well, including general web data, simulation data, and a very small amount of real-robot data detailed in Fig. \ref{fig:data}

\subsection{General Dataset}
\begin{figure}[h] 
\centering 
\includegraphics[width=1.0\linewidth]{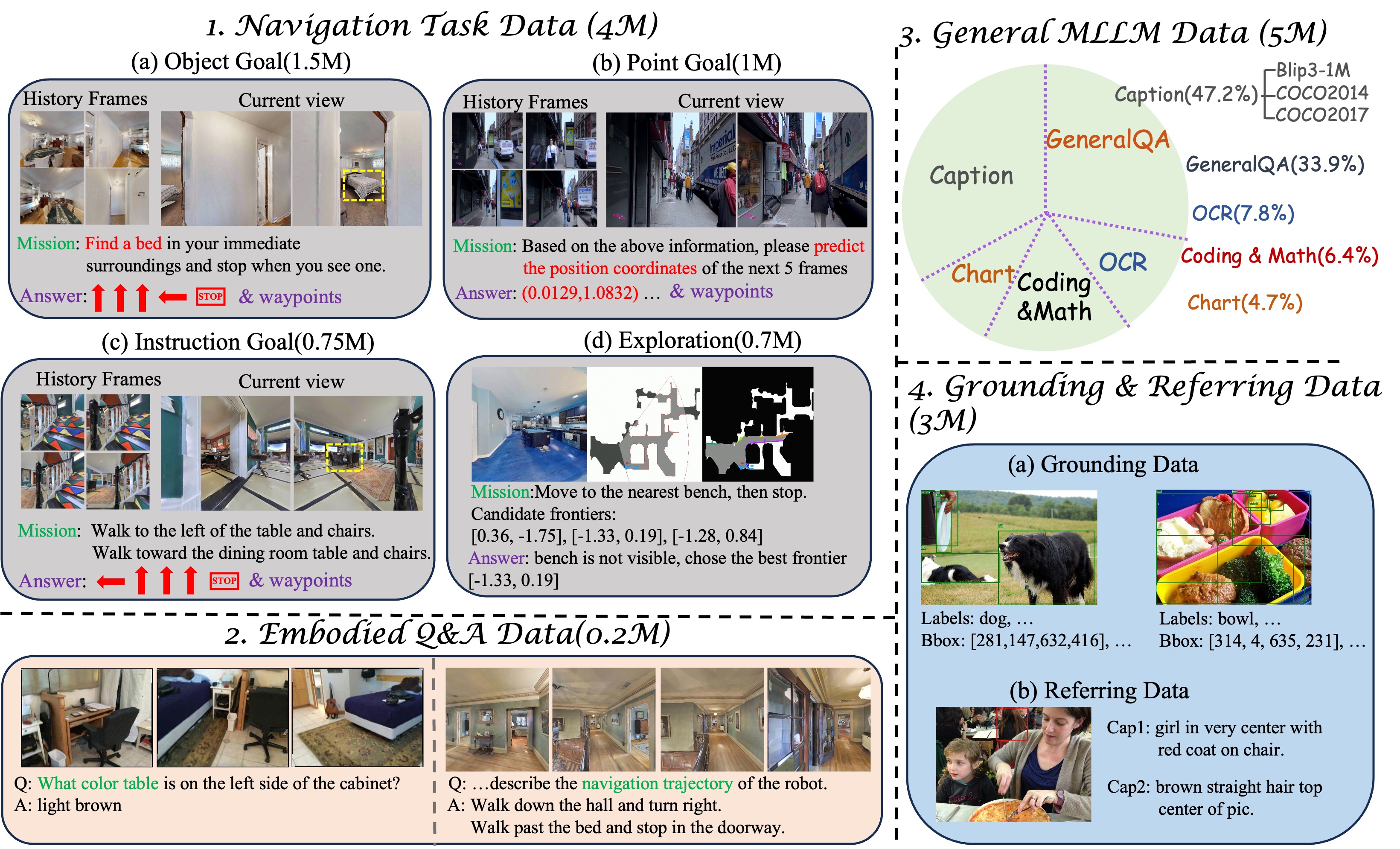}
\caption{Data composition overview. Four data types are used for training: Navigation task data, Embodied Q\&A data, General MLLM data and Grounding and referring data.}
\label{fig:data} 
\end{figure}

It is found in our experiment that models can learn the navigation paradigm relatively easily, whereas general-purpose capabilities remain challenging. To strengthen these abilities, we extended the training data with broad-coverage general-purpose datasets.
General-purpose datasets (general QA, image captioning, OCR, chart understanding, coding, and math) complement Vision-and-Language Navigation (VLN) by supplying foundational skills in language understanding, visual semantics, OCR recognition, structured reasoning, and algorithmic planning. These capabilities improve instruction comprehension, path planning, and overall robustness. They also introduce commonsense and functional priors, such as “bath towels are commonly found in bathrooms”. We draw on the open-source MAmmoTH-VL corpus~\citep{guo2024mammoth}, and subsampled examples to expand our general capabilities; the composition ratios are shown in Fig. \ref{fig:data}.

We further incorporated grounding and referring data to more reliably ground linguistic targets and relations to concrete pixels, instances, and locations in the scene, thereby yielding policies that are more interpretable, robust, and generalizable. This includes fine-grained language-to-target mapping—for example, precisely localizing language like “red sofa,” “door with a handle,” or “the second chair” to the correct image regions to support instance-level disambiguation. It also encompasses spatial and relational understanding, such as learning to ground spatial prepositions and relations (“to the left/in front of/at the corner”), and extracting actionable turning and stopping cues from relational descriptions like “the door next to the painting” or “after passing the hallway, turn right at the first intersection.” The referring and grounding data are sourced from RefCOCO series~\citep{chen2025revisiting} and Objects365~\citep{shao2019objects365}.

\subsection{Multi-Task Navigation Dataset}
\textbf{Point-goal data}. We use the open-source Citywalker corpus~\citep{liu2025citywalker}, which consists of first-person city-walking videos collected from public YouTube channels. Each trajectory comprises a single forward-view sequence. Long videos are segmented into 2-minute clips, then the camera pose for every frame are recovered by DPVO (Deep Patch Visual Odometry)~\citep{teed2023deep}.

\textbf{Instruct-goal data}. Under Habitat’s VLN-CE (continuous environment) setting and Matterport3D scens~\citep{chang2017matterport3d}, based on the public instruction–path pairs from R2R~\citep{anderson2018vision} and RxR~\citep{ku2020room}. For each trajectory, we store a panoramic sequence composed of three-views (front, left, right), the action and waypoint sequences, and the natural-language instruction.

\textbf{Object-goal data}. Specifically, open-vocabulary object navigation (OVON) data~\citep{yokoyama2024hm3d}. In Habitat-Matterport3D (HM3D)~\citep{ramakrishnan2021habitat} scenes, we randomly sample (start pose, target object category) pairs. The built-in shortest-path navigation algorithm is used to move the agent to the vicinity of the target. The same content are recorded as the instruct-goal data for each trajectory.

\textbf{Object-goal (with frontier-based exploration) data}. At each step the agent updates its occupancy map and visible region to identify all current frontiers. A policy then selects one frontier—favoring the shortest path while introducing limited randomness—as the current subgoal. Once the target object is found, the exploration episode terminates successfully, yielding a trajectory~\citep{zhu2025move}. Each trajectory record includes a single forward-view frame sequence, a unexplored frontiers sequence and a natural-language instruction.

\textbf{Embodied QA}. ScanQA~\citep{azuma2022scanqa} focuses on real-world indoor 3D scene understanding, with QA pairs centered on object locations, attributes, and spatial relations. R2R-EnvDrop~\citep{tan2019learning} addresses continuous visual navigation: scenes are from Matterport3D, trajectories are from R2R-CE, and the navigation instruction–trajectory pairs are recast into a QA format to strengthen alignment between linguistic expressions and visual observations.

\textbf{Navigation data Process}.Each action step corresponds to a 3D continuous pose, described jointly by position and orientation. Taking the agent’s current front-view coordinate frame as the origin, all other trajectory points are transformed into this local frame. Then, we project each 3D pose onto the ground plane, keeping only the planar coordinates $(x, y)$ and the heading angle $(\theta)$, and we additionally attach an ``arrival" flag to each step. In this way, each trajectory point is ultimately represented as a quadruple $(x, y, \theta, arrive)$, which serves as the primary target for waypoint optimization. Under this abstraction, since the underlying simulator ensure that any valid 2D path passing through a stair region can be realized as a 3D trajectory across floors, this representation naturally supports multi-floor navigation in HM3D: moving from one floor to another is encoded as following a sequence of 2D poses from the stair entrance to the stair exit, while the vertical motion is handled implicitly by the simulator. At the same time, the agent always receives 3D visual observations with full geometric information, so the stair structure and floor changes are reflected in the visual features.

\subsection{Discrete and Continuous Joint Training}

We adopt a two-stage training paradigm to balance language–vision semantics and continuous control. In Stage 1, we use an autoregressive (AR) objective to predict discrete variables (e.g., navigation action chunks, general-purpose semantic data, Embodied QA, grounding and referring data; see the four data types in Fig. \ref{fig:data}), to achieve alignment between language–vision and action. In Stage 2, we attach a flow-matching policy to the shared backbone to predict continuous waypoints, and perform joint training by including 20\% of the Stage-1 discrete data to prevent degradation of the base VLM during continuous-control fine-tuning. The continuous waypoint coordinates are normalized using min-max normalization to ensure stable training and better convergence.

A joint training scheme is particularly critical for Vision-and-Language Navigation (VLN), which requires strong general knowledge and semantic understanding, leading to substantial improvements in success rates in open environments.
Stage 1 is trained with 96 NVIDIA H20 GPUs for 120 hours, and Stage 2 is trained with 64 NVIDIA H20 GPUs for 48 hours with lower learning rate.

\section{Experiments}
\label{others}

\begin{table}[h]
\centering
\caption{Main comparison with prior methods on the Val-Unseen split of R2R-CE and RxR-CE.}
\label{tab:r2r and rxr}

\resizebox{\textwidth}{!}{
\begin{tabular}{lcccccccccccc}
\toprule
{\textbf{Method}} & \multicolumn{4}{c}{\textbf{Observation}} & \multicolumn{4}{c}{\textbf{R2R-CE Val-Unseen}} & \multicolumn{4}{c}{\textbf{RxR-CE Val-Unseen}} \\

  & S.RGB & Pano. & Depth & Odo. & NE$\downarrow$ & OS$\uparrow$ & SR$\uparrow$ & SPL$\uparrow$ & NE$\downarrow$ & SR$\uparrow$ & SPL$\uparrow$ \\

\midrule
HPN+DN*~\citep{krantz2021waypoint} & & \checkmark & \checkmark & \checkmark & 6.31 & 40.0 & 36.0 & 34.0 & - & - & - \\
CMA*~\citep{hong2022bridging} & & \checkmark & \checkmark & \checkmark & 6.20 & 52.0 & 41.0 & 36.0 & 8.76 & 26.5 & 22.1 \\
Sim2Sim*~\citep{krantz2022sim} & & \checkmark & \checkmark & \checkmark & 6.07 & 52.0 & 43.0 & 36.0 & - & - & - \\
GridMM*~\citep{wang2023gridmm} & & \checkmark & \checkmark & \checkmark & 5.11 & 61.0 & 49.0 & 41.0 & - & - & - \\
DreamWalker*~\citep{wang2023dreamwalker} & & \checkmark & \checkmark & \checkmark & 5.53 & 59.0 & 49.0 & 44.0 & - & - & - \\
Reborn*~\citep{an20221st} & & \checkmark &\checkmark & \checkmark & 5.40 & 57.0 & 50.0 & 46.0 & 5.98 & 48.6 & 42.0 \\
ETPNav*~\citep{an2024etpnav} & & \checkmark & \checkmark & \checkmark & 4.71 & 65.0 & 57.0 & 49.0 & 5.64 & 54.7 & 44.8 \\
HNR*~\citep{wang2024lookahead} & & \checkmark & \checkmark & \checkmark & 4.42 & 67.0 & 61.0 & 51.0 & 5.50 & 56.3 & 46.7 \\
\midrule
AG-CMTP~\citep{chen2021topological} & & \checkmark & \checkmark & \checkmark & 7.90 & 39.0 & 23.0 & 19.0 & - & - & - \\
R2R-CMTP~\citep{chen2021topological} & & \checkmark & \checkmark & \checkmark & 7.90 & 38.0 & 26.0 & 22.0 & - & - & - \\
InstructNav~\citep{long2024instructnav} & & \checkmark & \checkmark & \checkmark & 6.89 & - & 31.0 & 24.0 & - & - & - \\
LAW~\citep{raychaudhuri2021language} & \checkmark & & \checkmark & \checkmark & 6.83 & 44.0 & 35.0 & 31.0 & 10.90 & 8.0 & 8.0 \\
CM2~\citep{georgakis2022cross} & \checkmark & & \checkmark & \checkmark & 7.02 & 41.0 & 34.0 & 27.0 & - & - & - \\
WS-MGMap~\citep{chen2022weakly} & \checkmark & & \checkmark & \checkmark & 6.28 & 47.0 & 38.0 & 34.0 & - & - & - \\
AO-Planner~\citep{chen2025affordances} & & \checkmark & \checkmark & & 5.55 & 59.0 & 47.0 & 33.0 & 7.06 & 43.3 & 30.5 \\
Seq2Seq~\citep{krantz2020beyond} & \checkmark & & \checkmark & & 7.77 & 37.0 & 25.0 & 22.0 & 12.10 & 13.9 & 11.9 \\
CMA~\citep{krantz2020beyond} & \checkmark & & \checkmark & & 7.37 & 40.0 & 32.0 & 30.0 & - & - & - \\
NaVid~\citep{zhang2024navid} & \checkmark & & & & 5.47 & 49.0 & 37.0 & 35.0 & - & - & -  \\
Uni-NaVid~\citep{zhang2024uni} & \checkmark & & & & 5.58 & 53.5 & 47.0 & 42.7 & 6.24 & 48.7 & 40.9  \\
NaVILA~\citep{cheng2024navila} & \checkmark & & & & 5.22 & 62.5 & 54.0 & 49.0 & 6.77 & 49.3 & 44.0 \\
StreamVLN~\citep{wei2025streamvln} & \checkmark & & & & 4.98 & 64.2 & 56.9 & 51.9 & 6.22 & 52.9 & 46.0 \\
CorrectNav~\citep{yu2025correctnav} & \checkmark & & & & 4.24 & 67.5 & 65.1 & 62.3 & 4.09 & 69.3 & 63.3 \\
\midrule
OmniNav(w/o policy-head) & &\checkmark& & & 4.36 & 65.0 & 59.8 & 57.5 & 3.87& 64.1&53.9\\
\textbf{OmniNav} & &\checkmark& & & \textbf{3.74} & \textbf{74.6} & \textbf{69.5} & \textbf{66.1} & \textbf{3.77} & \textbf{73.6} & \textbf{62.0} \\
\bottomrule
\end{tabular}
}
\end{table}

\begin{table}[h]
  \centering
  \caption{Evaluation of object-goal navigation on HM3D-OVON, where * indicates OmniNav with slow thinking system.}
  \label{tab:ovon}
  \resizebox{\textwidth}{!}{\begin{tabular}{l|ccc|cc|cc|cc}
    \toprule
    \textbf{Method}& \multicolumn{3}{c|}{\textbf{Observation}} & \multicolumn{2}{c|}{\textbf{Val-Seen}} & \multicolumn{2}{c|}{\textbf{Val-Seen-Synonyms}} & \multicolumn{2}{c}{\textbf{Val-Unseen}} \\
    \cmidrule(lr){2-4} \cmidrule(lr){5-6} \cmidrule(lr){7-8}\cmidrule(lr){9-10}
    & \textbf{S.RGB}& \textbf{Depth}& \textbf{Odo.} & \textbf{SR}$\uparrow$ & \textbf{SPL}$\uparrow$ & \textbf{SR}$\uparrow$ & \textbf{SPL}$\uparrow$ & \textbf{SR}$\uparrow$ & \textbf{SPL}$\uparrow$ \\
    \midrule
    BC     &\checkmark & &  & 11.1 & 4.5  & 9.9  & 3.8  & 5.4  & 1.9  \\
    DAgger  &\checkmark & &  & 11.1 & 4.5  & 9.9  & 3.8  & 5.4  & 1.9  \\
    RL    &\checkmark & &   & 18.1 & 9.4  & 15.0 & 7.4  & 10.2 & 4.7  \\
    DAgRL  &\checkmark & &  & 41.3 & 21.2 & 29.4 & 14.4 & 18.3 & 7.9  \\
    BCRL   &\checkmark & &      & 39.2 & 18.7 & 27.8 & 11.7 & 18.6 & 7.5  \\
    VLFM*~\citep{yokoyama2024vlfm}  &\checkmark &\checkmark & \checkmark      & 35.2 & 18.6 & 32.4 & 17.3 & 35.2 & 19.6 \\
    DAgRL+OD~\citep{yokoyama2024hm3d}   &\checkmark &\checkmark &\checkmark  & 38.5 & 21.1 & 39.0 & 21.4 & 37.1 & 19.8 \\
    Uni-NaVid*~\citep{zhang2024uni}  &\checkmark & & & 41.3 & 21.1 & 43.9 & 21.8 & 39.5 & 19.8 \\
    MTU3D*~\citep{zhu2025move}   &\checkmark &\checkmark & \checkmark  & 55.0 & 23.6 & 45.0 & 14.7 & 40.8 & 12.1 \\
    \midrule
    OmniNav &\checkmark & & & 46.6 &23.3  & 50.4 &  28.5  & 43.5 & 27.3 \\  
    \textbf{OmniNav*(w/ cot)}&\checkmark &\checkmark & \checkmark& \textbf{56.1} & \textbf{30.0} & \textbf{68.6} & \textbf{38.8} & \textbf{59.2} & \textbf{33.2} \\
    \bottomrule
  \end{tabular}}
\end{table}

\textbf{Metrics} 
We evaluate navigation performance using success rate (SR), oracle success rate (OS), success weighted by path length (SPL), and navigation error (NE). Our evaluation protocol is consistent with prior work~\citep{zhang2024uni,zhu2025move} and follows standard practice.

\textbf{Instruct goal}
As shown in Table \ref{tab:r2r and rxr}, On the R2R-CE and RxR-CE benchmarks, we compare our model against all relevant competitors, including both discrete and continuous prediction methods. Notably, using only its fast system and pure RGB inputs, OmniNav achieves state-of-the-art success rates on both benchmarks. It surpasses the previous leading model by improving the success rate by 4.4\% on R2R-CE and 4.3\% on RxR-CE.

\textbf{Object goal}
To further validate open vocabulary generalization, we also compare OmniNav with prior methods on the HM3D-OVON benchmark. As shown in Table \ref{tab:ovon}, under purely visual inputs, OmniNav already surpasses the best existing approach by 2.7\%. However, given that the OVON task demands long-horizon and global planning, the limitations of a purely reactive fast system—such as getting trapped in local loops and exhibiting poor map coverage—become particularly pronounced. Therefore, we incorporate the slow system integrated with frontier-based reasoning, which necessitates the use of depth and odometry information to build and maintain an occupation map. This augmentation equips the agent with global spatial awareness and the capacity for proactive exploration, ultimately leading to superior overall performance(exceeding the strongest prior method by 18.4\%).



\textbf{Point goal}
We compare point-goal performance on the CityWalker benchmark, a state-of-the-art point-goal method that supports outdoor navigation scenarios. CityWalker adopts MAOE (Mean Average Orientation Error) as an open-set evaluation metric. On this open-set metric, our approach still outperforms the benchmark method (OmniNav:11.53\% vs CityWalker:15.23\%).

\textbf{Ablation Study}
The independent and synergistic contributions of the four key components policies are evaluated as show in Table.\ref{tab:r2r and rxr} and Table.\ref{tab:ablation}. 1) Fast system: autoregressively action chunks generation vs. continuous waypoints generation by flow-matching policy. On R2R-CE, RxR-CE, and OVON benchmarks, the degradation is substantial for action chunks. Their semantic tokens (e.g., left, right) align more easily with language, making them suitable for the first-stage training. However, because action chunks are coarse-grained motion control, continuous waypoints are better suited for fine-grained control.
2) Slow system (planning with frontier and long-term visual memory).
We primarily compare with/without the slow system on long-horizon exploration in OVON and find the largest improvements here. Once previously explored areas are recorded, the agent reduces redundant exploration and improves efficiency. Moreover, decomposing active exploration into subgoals (e.g., “go to the bedroom first”) and letting the fast system quickly approach each subgoal forms a hierarchical "plan–execute" loop, which better matches human reasoning and behavior in unfamiliar environments.
3) General data (general MLLM and referring/grounding). With the slow system enabled, adding the general-purpose datasets yields further stable gains.
4) CoT (explicit chain-of-thought outputs).
Using CoT makes the basis for subgoal selection in the slow system transparent, enabling process-level self-check and correction. It reduces cumulative errors in long chains and complex semantic tasks, producing stable improvements.
When all four are enabled, performance is best: the slow system provides semantically plausible long-horizon subgoals; the policy head executes with high precision and low latency; general data injects commonsense and language–vision alignment; and CoT provides auditable processes and self-correction.

\begin{table}[h]
\centering
\caption{Ablation study on HM3D-OVON Val-Unseen}
\label{tab:ablation}

\begin{tabular}{l|cccccc}  
\toprule
\textbf{Method} & \multicolumn{4}{c}{\textbf{Module}}  & \multicolumn{2}{c}{\textbf{Val-Unseen}} \\
\cmidrule(lr){2-5} \cmidrule(lr){6-7}
 & policy-head & slow-system & general data & COT &  SR$\uparrow$ & SPL$\uparrow$\\ 
\midrule
{OmniNav} &            &            &            &            & 35.3 & 22.1 \\
                         & \checkmark &            &            &            & 43.5 & 27.3 \\
\cmidrule(lr){1-7} 
{OmniNav*}& \checkmark & \checkmark &            &            & 55.9 & 30.7 \\
                         & \checkmark & \checkmark & \checkmark &            & 57.7 & 32.9 \\   
                         & \checkmark & \checkmark & \checkmark & \checkmark & \textbf{59.2} & \textbf{33.2} \\
\bottomrule
\end{tabular}
\end{table}

\begin{figure}[h] 
\centering 
\includegraphics[width=1.0\linewidth]{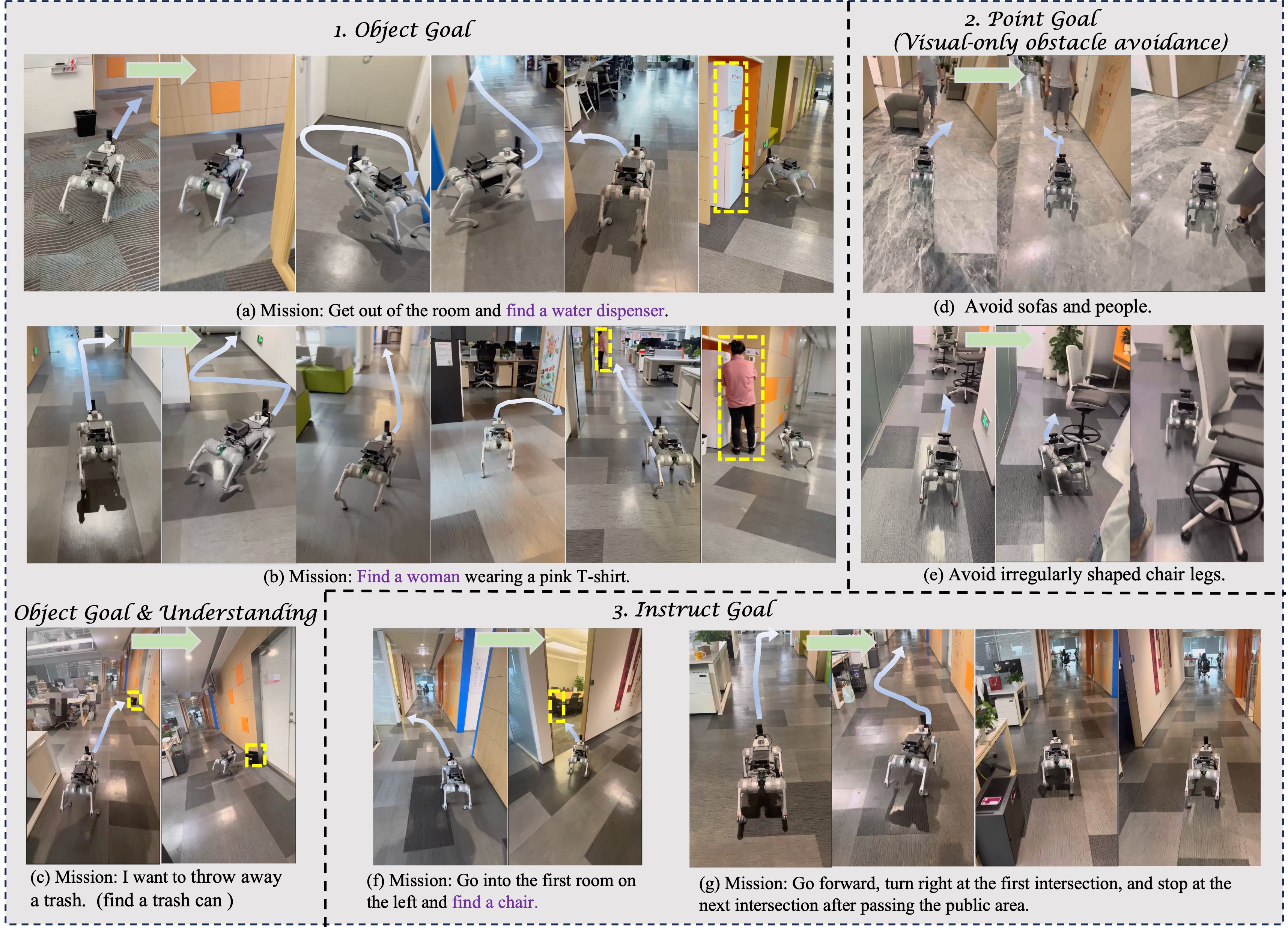}
\caption{Real-world deployment. It shows third-person view of the three different navigation tasks which are deployed in a zero-shot setting. The gradient blue arrows indicate the trajectory, and the yellow box marks the target location. Our model demonstrates highly effective navigation performance on the real quadruped robot.}
\label{fig:deploy} 
\end{figure}

\textbf{Real-world deployment}
In real-robot deployment see Fig. \ref{fig:deploy}, we deploy the fast system component of the model architecture—comprising the VLM and the policy head on a cloud server with an RTX 3090 GPU. The history buffer holds up to 20 front-view frames with at a resolution of 120×106, while the current input is tri-view at 480×426. The system runs at over 5 Hz. For all tasks, it outputs waypoints, which are then fed into the onboard speed control module, which, based on the current speed and maximum acceleration, generates a set of candidate speeds. It then selects the optimal speed to approach the waypoint (selection criteria: minimize speed changes to maintain high speed as much as possible, and be closest to the target waypoint). 

Deploying the full slow system in real-world settings requires additional engineering, such as robust real-time integration with LiDAR/depth estimation. This paper mainly focuses on validating the effectiveness of the dual-system collaboration framework in terms of navigation performance and behavior. A full physical deployment of the complete slow system, and systematic optimization of its performance under real-world constraints (including detailed latency–frequency trade-off analysis), constitutes an important avenue for future research.

\section{CONCLUSION \& FUTURE WORKS}

\textbf{Conclusion}
The core of OmniNav is a fast–slow dual-system architecture: the fast system, conditioned on VLM-fused multimodal context, employs a flow-matching policy to generate future continuous waypoints, achieving low latency, and high-precision closed-loop control; the slow system plans subgoals and subtasks supported by long-horizon visual memory and frontiers, and introduces explicit CoT for interpretability and self-correction. This architecture supports most basic tasks in embodied navigation. Through unified multimodal tokenization, different tasks (instruct goal, object goal, point goal) are seamlessly handled within a single model.
On the training side, we adopt a two-stage scheme, where the second stage’s joint training of discrete and continuous values prevents continuous-control fine-tuning from eroding the base VLM’s capabilities—an approach that can be broadly applicable. We also incorporate sizable general-purpose data and referring/grounding data to bolster language understanding, visual semantics, structured reasoning, and commonsense priors for VLN, thereby improving generalization and robustness in embodied navigation.
Experimentally, OmniNav improves success rates over the current best on R2R-CE and on RxR-CE, achieves the best performance on OVON, and benefits further from the slow-system design. Real-world quadruped robot deployment demonstrates the engineering feasibility of up to 5 Hz cloud inference with tri-view inputs and a 20-frame history buffer.
Overall, the high spatial precision and low latency of continuous waypoints, the unified multimodal interface, the fast–slow system collaboration, and joint training collectively underpin OmniNav’s strong performance across benchmarks, evidencing solid open-set generalization and practical deployment potential.

\textbf{Future Works}
We aim to develop a more semantics-driven, learning-based subgoal selection strategy and realize purely visual memory capabilities—for example, remembering the semantic regions already explored. In addition, we plan to build a retrievable, lifelong spatiotemporal memory to enable lifelong navigation.


\bibliography{iclr2026_conference}
\bibliographystyle{iclr2026_conference}


\clearpage 

\appendix 

\section{Appendix} 

\subsection{Analysis of Training Data and Model scale}
\textbf{Analysis of Training Data} The three additional data components shown in Figure \ref{fig:data} are ablated individually. A qualitative analysis of specific examples reveals the following: 1) Removing Embodied Q\&A or Grounding/referring noticeably degrades performance on small background objects (e.g., ``picture", ``flowerpot"), suggesting these data improve recognition of small objects. 2) Removing General MLLM data leads to failures on irregular objects (e.g., ``handrail", ``stair"), implying general vision–language data helps with such targets. These findings motivate constructing specific datasets to address failure modes. For example, the model still struggles with ``carpet" and ``clothes" where complex folds and textures are involved; we therefore can curate dedicated VQA-style data emphasizing these patterns to strengthen fine-grained visual reasoning on texture-heavy objects.

\begin{table}[h] 
    \centering
    \caption{Ablation Study of Data on HM3D-OVON Val-Unseen}
    \label{tab:quantization_results}
    \begin{tabular}{cccc}
        \toprule
        {Embodied Q\&A Data} & {Grounding and Referring Data} & {General MLLM Data} &{Ovon-Unseen}  \\
        \midrule
           &  &  & 55.9 \\
                        & \checkmark & \checkmark & 56.5 \\
                       \checkmark &  & \checkmark & 56.7 \\
                       \checkmark & \checkmark &  & 57.0 \\
                       \checkmark & \checkmark & \checkmark & 57.7 \\
        \bottomrule
    \end{tabular}
\end{table}

\textbf{Analysis of Model Scale} The influence of model size (Embodied Q\&A, Grounding and Referring, and General MLLM data) is also analyzed. When including these additional data, the 3B and 7B models exhibit nearly identical navigation performance, indicating that once such data are incorporated, simply scaling up the model size brings little further improvement. 

In contrast, in the absence of this additional data, a noticeable performance gap emerges, with the 7B model outperforming the 3B model. Our analysis suggests two key factors: 1) the performance of the 3B model appears to be constrained by data sufficiency rather than its inherent capacity. When provided with diverse and abundant data, its performance becomes comparable to that of the 7B model, indicating that model size itself is not the primary bottleneck in this data-rich scenario. 2) based on an analysis of failure cases, we find that further performance gains are limited by the intrinsic difficulty of the task: regardless of model size, the recognition of complex objects such as clothes and mirrors remains unstable. 

Beyond these two ablation study, a more systematic study of scaling laws and optimal training configurations—e.g., how data quality, data composition, and more model size jointly affect performance—would also be highly valuable. Due to computational limits we have not yet conducted such a systems-level exploration, and we view this as an important direction for future work.

\begin{table}[h] 
    \centering
    \caption{Ablation Study of Model Size on HM3D-OVON Val-Unseen}
    \label{tab:quantization_results}

    \begin{tabular}{ccc}
        \toprule
        {Model} & {Additional Data} & {Ovon-Unseen} \\
        \midrule 
                       3B &  & 55.9 \\
                       7B  &  & 57.2 \\
                       3B  & \checkmark & 57.7 \\
                       7B & \checkmark & 57.9 \\
        \bottomrule
    \end{tabular}
\end{table}

\end{document}